\documentclass[runningheads]{llncs}
\usepackage{graphicx}
\usepackage{times}
\usepackage{helvet}
\usepackage{courier}
\usepackage{booktabs}
\usepackage{hyperref}
\usepackage{algpseudocode}
\usepackage{algorithm}
\usepackage{graphicx}
\usepackage[labelfont=bf]{caption}
\usepackage[utf8]{inputenc}
\usepackage{amsmath}
\usepackage{multicol}

\begin{document}
\title{Adaptive Tabu Dropout for Regularization of Deep Neural Networks}
%
%

\author{
Md. Tarek Hasan$^1$, Arifa Akter$^1$, Mohammad Nazmush Shamael$^1$, Md Al Emran Hossain$^1$, H. M. Mutasim Billah$^1$, Sumayra Islam$^1$, and Swakkhar Shatabda$^2$\\ $^1$Department of Computer Science and Engineering, United International University\\$^2$Department of Computer Science and Engineering, Brac University\\tarek@cse.uiu.ac.bd, \{aakter181254, mshamael181062, mhossain181144, hbillah181290, sislam181123\}@bscse.uiu.ac.bd, swakkhar.shatabda@bracu.ac.bd
}
\authorrunning{Hasan et al.
}
%
\institute{
}
\maketitle              
\begin{abstract}
Dropout is an effective strategy for the regularization of deep neural networks. Applying tabu to the units that have been dropped in the recent epoch and retaining them for training ensures diversification in dropout. In this paper, we improve the Tabu Dropout mechanism for training deep neural networks in two ways. Firstly, we propose to use \textit{tabu tenure}, or the number of epochs a particular unit will not be dropped. Different tabu tenures provide diversification to boost the training of deep neural networks based on the search landscape. Secondly, we propose an adaptive tabu algorithm that automatically selects the tabu tenure based on the training performances through epochs. On several standard benchmark datasets, the experimental results show that the adaptive tabu dropout and tabu tenure dropout diversify and perform significantly better compared to the standard dropout and basic tabu dropout mechanisms.

\end{abstract}
\section{Introduction}
Deep neural networks are a powerful machine learning system that can obtain very effective results in many applications such as natural language processing, bioinformatics, computer vision, and other similar fields. Deep neural networks containing a large number of parameters and multiple non-linear hidden layers can grasp many intricate relations in the data. However, while trying to understand these intricate relations, the model tends to perfectly fit the training data. In order to overcome this problem dropout can be used. Dropout \cite{hinton2012improving,krizhevsky2012imagenet} is an effective regularization technique that is designed to tackle the overfitting problem in deep neural networks. During the training phase, we close some of the neurons in the network for each epoch. This allows us to construct a ‘thinned’ network for each epoch. The final model is a combination of these ‘thinned’ models. This method produces models with superior generalization for the test data.

Although dropout can provide better results to reduce overfitting, the methods used to determine which neurons get dropped can affect the generalization of test data greatly. To resolve this problem, multiple approaches to dropout have been proposed over the years with varying levels of effectiveness. The overlap and the difference of neurons between two epochs can affect the generalization of the data. Tabu dropout \cite{ma2020dropout} tries to control this overlap and difference by not allowing a neuron to be dropped twice in a row. In tabu dropout, only the status of the neurons of the previous forward propagation needs to be stored. This allows the dropout method to increase the diversification of the neural network while being computationally efficient. In practice, Tabu Dropout outperforms many dropout techniques like AlphaDropout~\cite{klambauer2017self}, Curriculum dropout~\cite{morerio2017curriculum} and standard dropout. However, further control of the overlap and difference of neurons between two epochs can be achieved by controlling how many epochs a single neuron is not allowed to be dropped after being dropped once. Thus, increasing the generalization for the test data.

In this paper, we propose two methods to further improve the tabu dropout method. The first method is a regularization technique that controls how many epochs a single neuron is prohibited from being dropped after being dropped once. We call this method Tabu Tenure Dropout. The Tabu Tenure method increases the diversity of a neural network model while reducing the error rate. It allows us to have additional control over the overlap and difference of neurons between two epochs. The second method is an Adaptive Tabu Dropout algorithm. Using this algorithm, we dynamically select which Tabu Tenure to use during the training stage to get the optimum results. This algorithm allows us to harness the full potential of the Tabu Tenure Dropout by adaptively choosing the best-suited Tabu Tenure for the train data. We conducted our experiments on various standard datasets like MNIST, Fashion MNIST, CIFAR-10 and CIFAR-100 and got promising results. All of the Tabu Tenure dropout algorithms tested performed better than the standard Tabu dropout. Amongst the Tabu Tenure dropouts tested, Tabu Tenure 6 produced the best results in all experiments except the MNIST-CNN1 experiment. Using the Tabu Tenure dropout algorithm we reduced the error rate up to 62.06\% when compared with the standard tabu dropout during our testing. The second method called Adaptive Tabu dropout is formulated as a multi-armed bandit problem where we calculated the reward based on the loss per epoch and arms are the tabu tenure($TT$) values. We tested multiple policies to the Adaptive tabu tenures selection like a random selection, greedy selection, epsilon ($\epsilon$) greedy selection, probabilistic selection, and Softmax probabilistic selection. The Adaptive Tabu Dropout achieved similar or improved accuracy and error rate in general when compared to the Tabu Tenure dropout.

\section{Related Work}

Ma et al. \cite{ma2020dropout} presented a new dropout diversification approach that seeks to generate a more diverse neural network topology in fewer iterations. No neuron can drop twice in a row. The AS-Dropout \cite{chen2021adaptive} model drops most of the neurons based on the neurons’ activation functions. Though the model controls the proportion of the active neurons, it cannot accurately control it at a particular value. So, the dropout technique learns both confidence and uncertainty. The contextual dropout \cite{osti_10273825} is an efficient alternative to data-independent dropouts, which learns the dropout probabilities. It is well suited for both the Bernoulli dropout and the Gaussian dropout. The model may be used on a wide range of models with just a minor increase in memory and computational cost.

In contrast to multi-layer perceptrons, Irsoy and Alpaydın  \cite{irsoy2021dropout} offer a form of dropout for hierarchical mixtures of experts that are true to the tree hierarchy described by the model, rather than a flat, unit-wise independent application of dropout. The model is overfitted only if the number of levels and leaves is too high. AutoDropout \cite{pham2021autodropout} finds the dropout patterns efficiently for both image recognition as well as language modeling. A controller learns to produce a dropout pattern at every channel and layer of a target network. The dropout pattern is then utilized to train the target network, and the validation performance is then utilized as a learning signal for the controller. Li et al. \cite{li2021adaptive} presented an adaptive dropout method in which neural network and Variational Auto-encoder (VAE) are used alternately in the training phase. To regularize its hidden neurons, the model adaptively sets activity to zero.

As the traditional binary dropout method is not that precise, Tang et al. \cite{tang2020beyond} have proposed to merge distortions onto feature maps by utilizing the Rademacher complexity.  Using the generalization error bound, randomly selected elements in the feature maps are changed with particular values during the training phase. Hu et al. \cite{hu2020surrogate} have presented a simple and efficient surrogate dropout as an alternative to the learning parameter in the Bernoulli distribution. It learns the parameters by using concrete distribution. There are two steps of the model technique. For measuring the significance of each neuron, the initial step trains a surrogate module that may be improved alongside the neural network. When the network converges, the surrogate module’s output is used as a guiding signal for removing particular neurons, approximating the ideal pre-neuron drop rate.

\section{Proposed Method}

The tabu technique \cite{glover1989tabu} is commonly used in local search algorithms, and it uses a memory structure (referred to as the tabu list) to prevent the local search from returning a previously visited candidate solution. In \cite{ma2020dropout}, the authors have presented a new dropout technique based on the tabu strategy named Tabu Dropout. Algorithm \ref{Algo:Standard Tabu Dropout} shows how the forward propagation works with the Tabu Dropout technique described by \cite{ma2020dropout}. It prevents a neuron from being dropped in a consecutive epoch. In the rest of the section, we present two improvements of the standard tabu dropout: Tabu Tenure Dropout that generalizes the standard Tabu Dropout by extending it using a \textit{tabu tenure} and an Adaptive Tabu Dropout that can select the tabu tenure automatically based on the learner's performance.

Algorithm \ref{Algo:Standard Tabu Dropout} shows the standard Tabu Dropout Algorithm, where $x$ represents a hidden layer of neural network architecture. Initially, Tabu is set to none and $p$ is the dropout rate. At the time of training, the mask is created using the Bernoulli Distribution that represents which neurons should be dropped in this epoch. If nothing is set to Tabu, we do not have anything to compare with. But if the tabu is not none, then we need to check the neuron we are planning to drop whether it has already dropped in the previous epoch or not. If it has already dropped in the previous epoch then we would not consider dropping this neuron in this epoch also. After that, by doing the Hadamard product of mask and $x$ the value of $x$ will be updated and normalized afterward. The mask value should be set to \textit{tabu} to store the current state so that we can check whether a neuron has already dropped in the previous epoch or not.

\subsection{Tabu Tenure Dropout}

In this work, we propose a modified Tabu Dropout technique and named it Tabu Tenure Dropout. We generalize the idea of tabu not only for the consecutive epoch but for a certain number of epochs. We call this \textit{tabu tenure} denoted by $TT$. In other words, a neuron cannot be dropped if it has been dropped in previous $TT$ epochs. It thus results in increased diversity in the dropout of units. In the case of standard Tabu Dropout proposed in \cite{ma2020dropout}, the value of $TT$ is 1.

\begin{multicols}{2}
\begin{algorithm}[H]
\caption{Standard Tabu Dropout}
\label{Algo:Standard Tabu Dropout}
\begin{algorithmic}[1]
\State {x : a hidden layer of neural network architecture}
\State $Tabu \gets \textbf{None}$
\State $ p \gets $ {Dropout Rate}
\\
\Procedure{Tabu\_DROPOUT}{$x$}
\If {not training}
\State return x
\EndIf
\State {ones} $  \gets $ {number of ones of size}
\State {mask} $ \gets $ {Bernoulli(ones $\odot$ p)}
\If {{$Tabu$} $\neq$ $\textbf{None}$}
\State {mask} $ \gets $ {mask $\parallel$  ({$Tabu$}  $\land$ ones)}
\EndIf
\State {x} $\gets$ {mask $\odot$ x}
\State {x} $\gets$ x / (1 - $p$)
\State {$Tabu \gets$} mask 
\State return x
\EndProcedure
\end{algorithmic}
\end{algorithm}
\columnbreak

\begin{algorithm}[H]
\caption{Tabu Tenure Dropout}
\label{Algo:Tabu Tenure Dropout}
\begin{algorithmic}[1]
\State {x : a hidden layer of neural network architecture}
\State $Tabu \gets \textbf{None}$
\State $ p \gets${Dropout Rate}
\State $ epoch \gets 0$
\State $TT \gets $ {Tabu Tenure Value}
\\
\Procedure{Tabu\_Tenure\_DROPOUT}{$x$}
\If {not training}
\State return x
\EndIf
\State {ones} $  \gets $ {number of ones of size}
\State {mask} $ \gets $ {Bernoulli(ones $\odot$ p)}

\If {{$Tabu$} $\neq$ $\textbf{None}$}
\State {mask} $ \gets $ {mask} $\parallel$ (({$Tabu$ -- $epoch$} $\leq$ {$TT$}) \& ({$Tabu$} $\neq$ 0))
\EndIf

\State {x} $\gets$ {mask $\odot$ x}
\State {x} $\gets$ x / (1 - $p$)
\State Flip mask
\State {mask} $ \gets $ {mask} $\odot$ {$epoch$}
\State {$Tabu \gets$} maximum(mask, {$Tabu$}) 
\State {$epoch \gets$} {$epoch + 1$}

\State return x
\EndProcedure
\end{algorithmic}
\end{algorithm}

\end{multicols}

Algorithm \ref{Algo:Tabu Tenure Dropout} shows how the forward propagation works with the Tabu Tenure Dropout. It tracks the epoch number and checks whether the neuron has already been dropped in the previous $TT$ epochs, if it has already been dropped then it can not be dropped in the current epoch. This is how we can increase the diversification of the neural network architecture in a smaller number of epochs.

Algorithm \ref{Algo:Tabu Tenure Dropout} shows the Tabu Tenure Dropout where $x$ represents a hidden layer of neural network architecture. Like Algorithm \ref{Algo:Standard Tabu Dropout}, here \textit{Tabu} is set to none and $p$ is the dropout rate. As the Tabu Tenure can consider more than one epoch while dropping a neuron that is why we need to track the epoch number and initially it’s set to zero. $TT$ is the Tabu Tenure Value which means while dropping a neuron how many previous epochs we need to check whether that neuron has already dropped in previous $TT$ epochs or not.

At the time of training, we create mask using Bernoulli distribution similar to Algorithm \ref{Algo:Standard Tabu Dropout}. If the \textit{tabu} is not none, we need to check whether the neuron we are considering to drop has already dropped in the previous $TT$ epochs or not. After that we update the value of $x$ by Hadamard product with mask. Next, we flip the mask value and again Hadamard product with the current value of epoch. Now the mask represents the neurons those are dropped in the current epoch are set as epoch number and those have not dropped in the current epoch set as zero.  Now the \textit{tabu} will be updated by the element-wise maximum operation which represents the recent epoch where the neuron has dropped. After that, we increase the epoch value by 1. The only difference between Algorithm \ref{Algo:Standard Tabu Dropout} and Algorithm \ref{Algo:Tabu Tenure Dropout} is Algorithm \ref{Algo:Standard Tabu Dropout} considers only previous epoch while dropping a neuron. On the contrary, Algorithm \ref{Algo:Tabu Tenure Dropout} consider $TT$ number of epochs while dropping a neuron where the value of $TT$ is fixed before starting the training phase.

\subsection{Adaptive Tabu Dropout}

In Tabu Tenure Dropout, we consider $TT$ previous epochs while dropping a neuron in hidden layers. Instead of using a fixed number of $TT$, we can change the $TT$ value several times during training and make the algorithm to learn and set the $TT$ value in an adaptive manner. Here we formulate Adaptive Tabu Dropout as a multi-armed bandit problem. The arms or actions of the bandit are the $TT$ values taken from a set $\mathcal{T}$ containing several $TT$ values for selection and the reward, $R_i$ in epoch $i$ will be decided based on the loss function. 
The value function $Q_t(a)$ at epoch $t$ value of an arm $a$ will be estimated an average or expected value of the rewards when an arm $a$ is selected prior to epoch $t$, in our case an arm is the $TT$ values selected. In this paper, we have considered a \textit{sample-average} technique for the value function $Q_t(a)$. The equation for calculating  $Q_t(a)$ is given in Equation~\ref{formula:Q}.
\begin{equation}
    Q_t(a) = \frac{\sum_{i=1}^{t-1} R_i\cdot\phi(a) }{\sum_{i=1}^{t-1}\phi(a)}
    \label{formula:Q}
\end{equation}

Here, $\phi(a)$ is a function that returns 1 if $TT$ value $a$ is selected in epoch $i$ or 0 if it is not selected and $R_i$ is the reward returned by the selected $TT$ value in epoch $i$. The default values of $Q_t(a)$ is set to 0 is case any action is yet to be selected.

\subsubsection{Tabu Tenure Selection}
There are several techniques for choosing arms in each epoch, for example: random, greedy, epsilon greedy, probabilistic, Softmax probabilistic, and many others \cite{sutton2018reinforcement}.
For adaptive tabu dropout, we propose to change the $TT$ value after specific number of epochs in the training. Thus we introduce a new hyper-parameter for the algorithm named \textit{adaption period}. After each \textit{adaption period}, a new $TT$ value is selected based on the $TT$ selection strategy based on the value function $Q_t(a)$. Our adaptive tabu selection algorithm uses a policy, $\pi(a)$ to select the values of $TT$. The policy function $\pi(a)$ specifies a probability distribution over the possible $TT$ values in the set $\mathcal{T}$. In our experiments, we used the $TT$ values in the range $1,\dots,TT_{max}$. In this paper we have experimented with five different policies. They are described in the following.

\paragraph{Random Policy} The first policy is a exploration based strategy that selects a $TT$ value from the set $\mathcal{T}$ using a uniform random distribution. The policy is defined in the following equation.
\begin{equation}
\pi(a) = \frac{1}{|\mathcal{T}|} \text{,        }\forall a \in \mathcal{T}
\label{formula:random} 
\end{equation}

\paragraph{Greedy Policy} In greedy technique, we will choose the arm or $TT$ value with the largest $Q_t(a)$ denoted $TT^*$ defined in the following equation:

\begin{equation}
    TT^*= arg\max_a Q_t(a) \text{ (with ties broken arbitrarily)} 
\end{equation}

The greedy policy which is an exploitation based strategy is given in the following equation:
\begin{equation}
\pi(a) = \begin{cases}
1, \text{ if } a = TT^*\\
0, \text{ if } a \neq TT^*
\end{cases}
\label{formula:greedy}
\end{equation}

\begin{figure*}[b]
    \centering
\begin{tabular}{cccc}
  \includegraphics[width=0.25\linewidth]{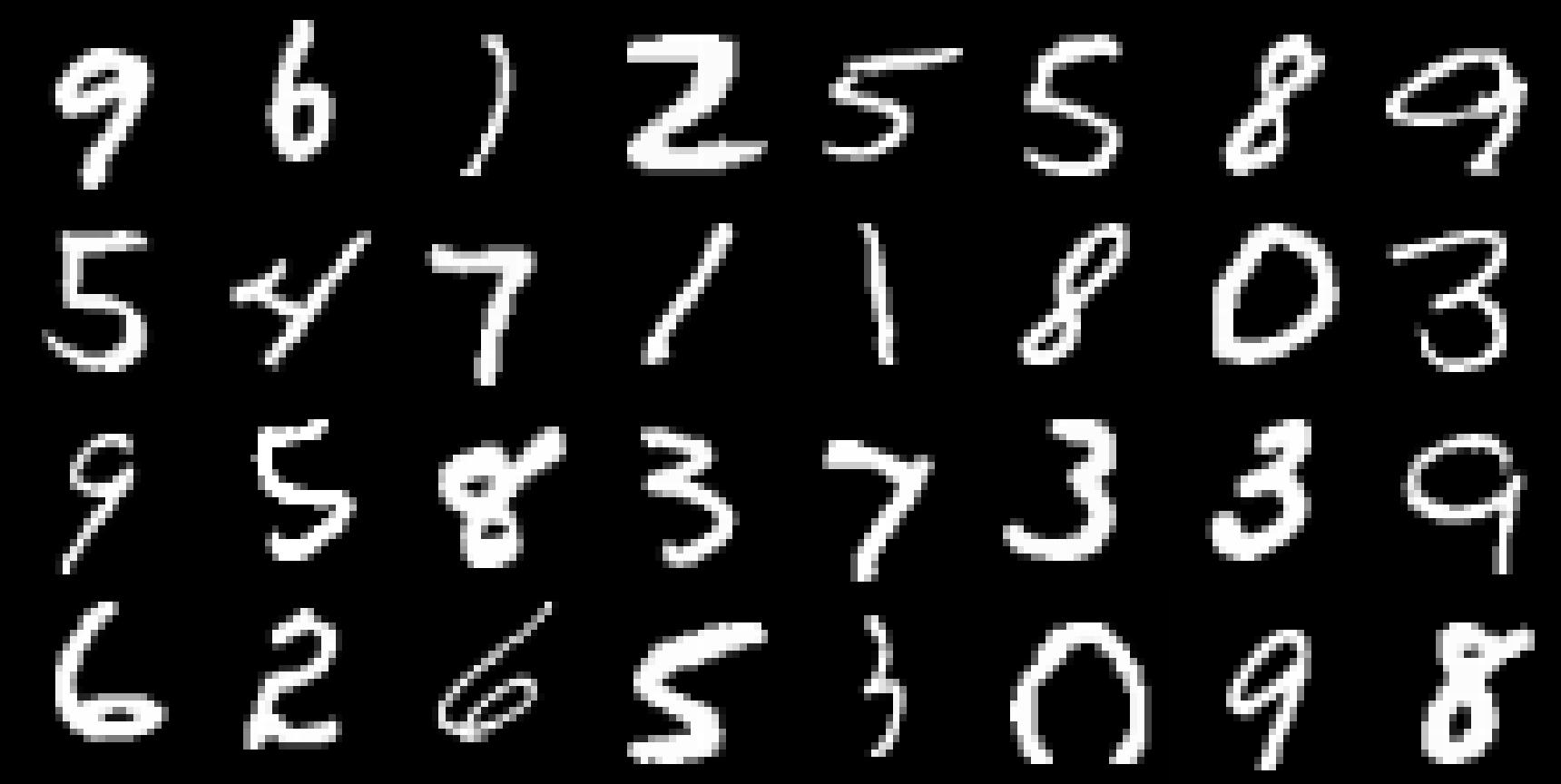} &  \includegraphics[width=0.25\linewidth]{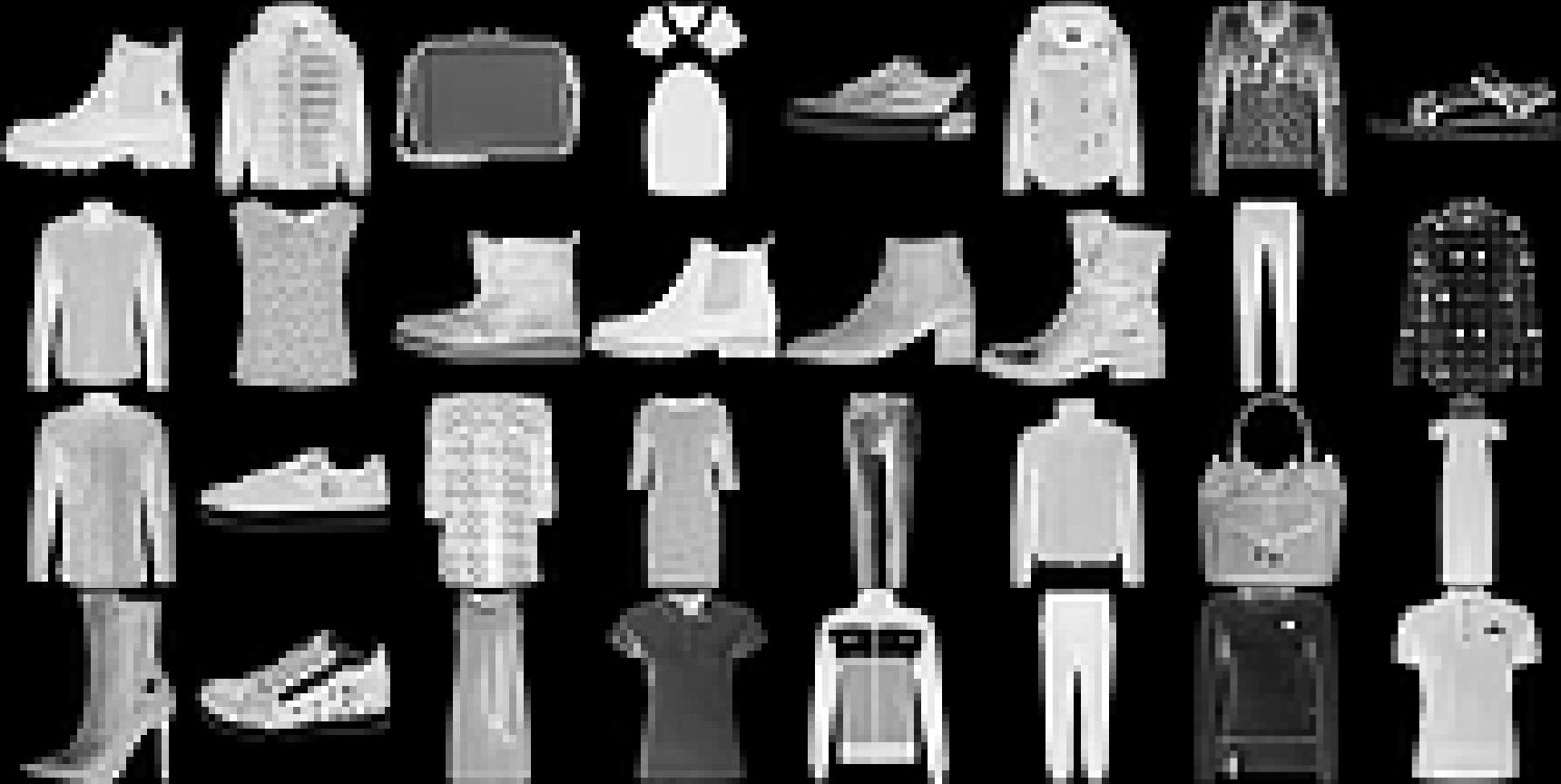}&
  \includegraphics[width=0.25\linewidth]{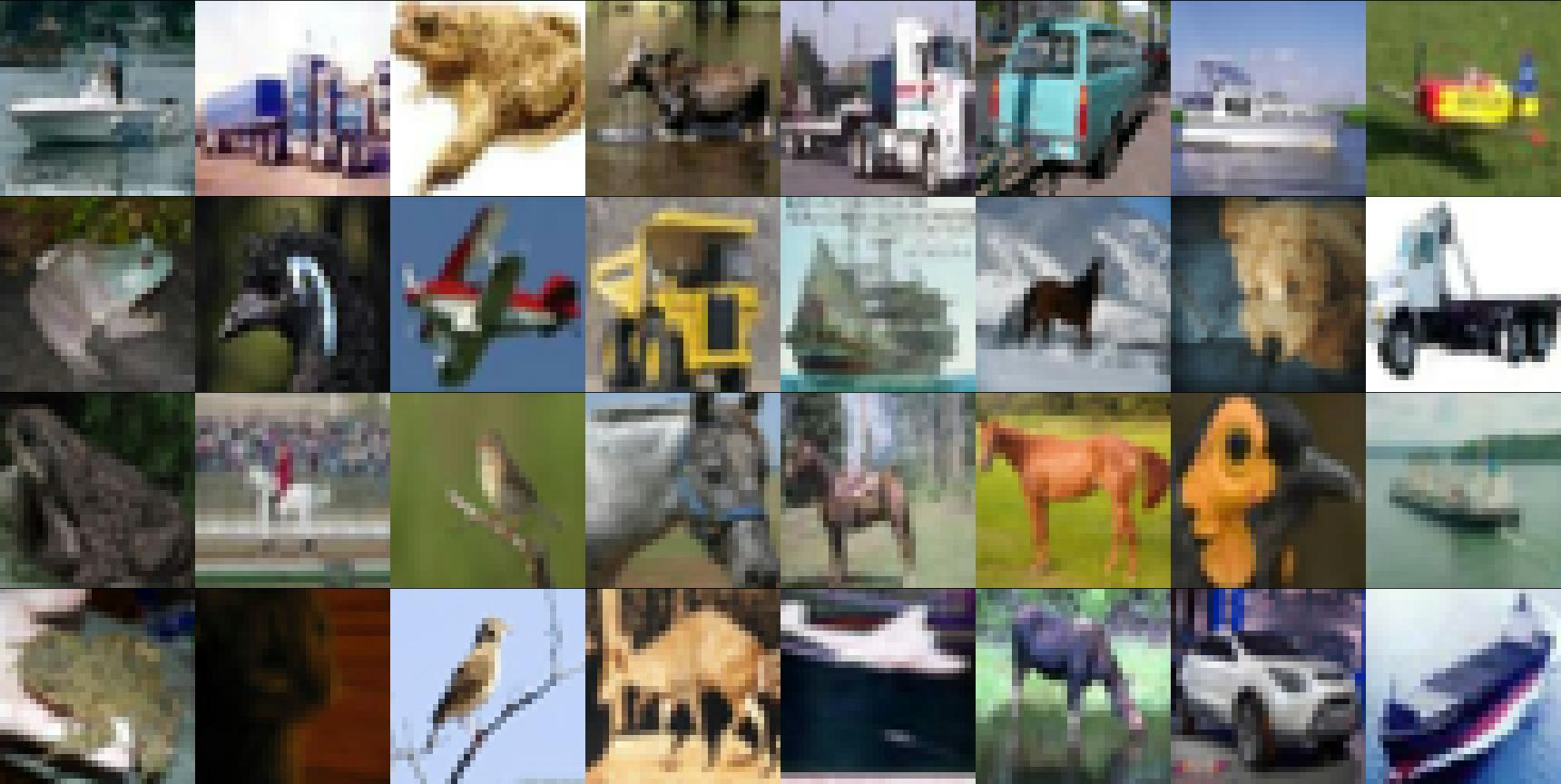} & 
  \includegraphics[width=0.25\linewidth]{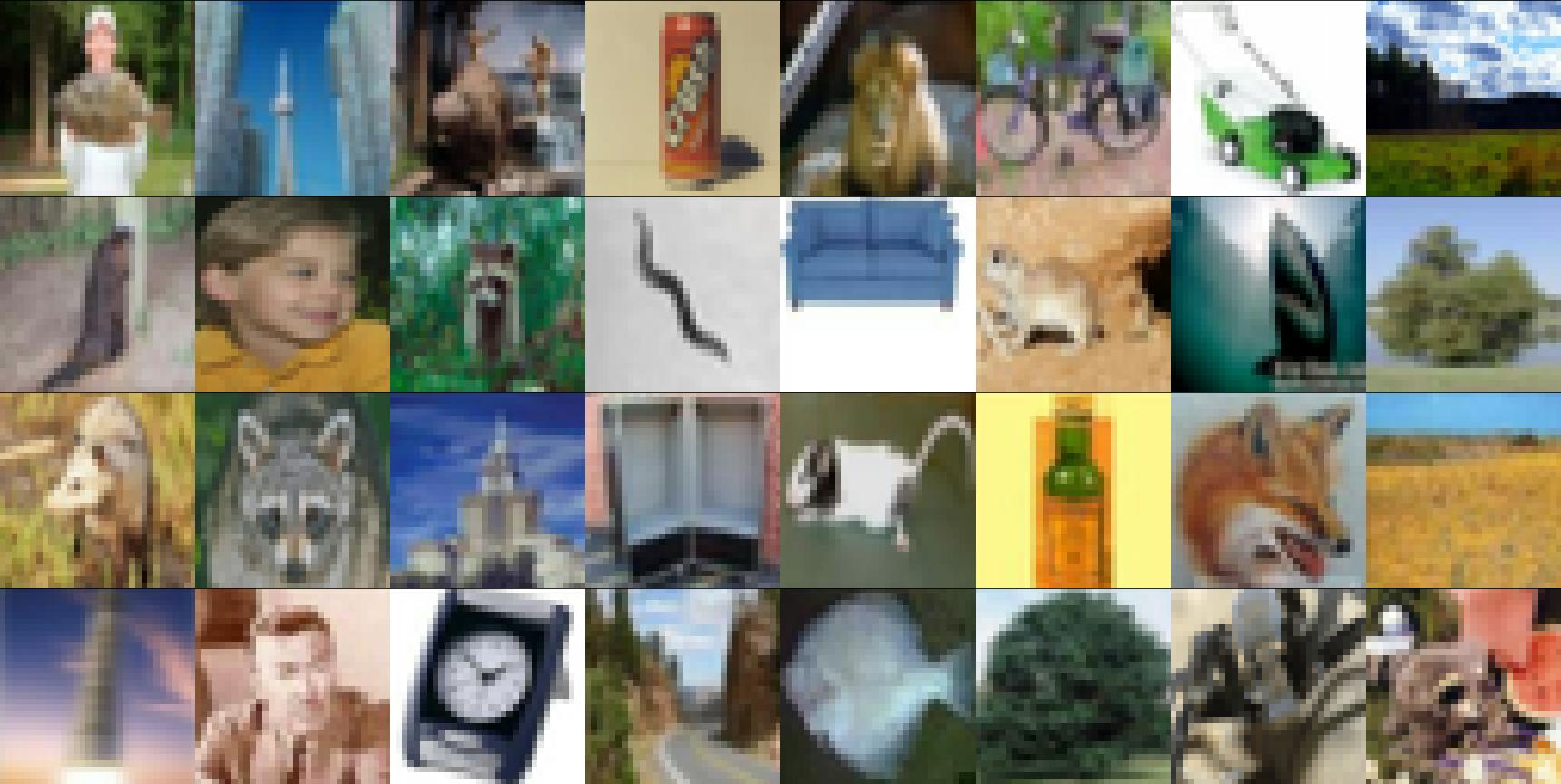}\\
  (a) & (b)&(c) & (d)\\
  
\end{tabular}

    \caption{Samples from different datasets (a) MNIST, (b) Fashion-MNIST, (c) CIFAR-10 and (d) CIFAR-100. }
    \label{fig:datasets}
\end{figure*}

\paragraph{Epsilon greedy Policy}For balancing between exploration and exploitation, $\epsilon$-greedy policy is used. Here, $\epsilon$ is the exploration parameter taking values from the range $[0,1]$.
The higher the value of epsilon the algorithm gives more emphasis to exploration and lower the value is it gives more emphasis to exploitation. A highest value of 1 makes the policy random and a lowest value 0 makes the algorithm greedy. In our experiments, we have set the value of epsilon, $\epsilon=0$.5.   
The $\epsilon$-greedy policy is defined in the following equation:

\begin{equation}
\pi(a) =\begin{cases}
1-\epsilon + \frac{\epsilon}{|\mathcal{T}|}, \textbf{  if  } a = TT^*\\
\frac{\epsilon}{|\mathcal{T}|}, \textbf{  if  } a \neq TT^*
\end{cases}
\end{equation}

\paragraph{Probabilistic Policy} We also use a policy that learns a probability distribution based on the $Q_t(a)$ values. Here the $Q_t(a)$ values are normalized to convert into probability. The policy is defined in the following equation:


\begin{equation}
\pi(a) = \frac{Q_t(a)}{\sum_{a\in\mathcal{T}}Q_t(a)} \text{,        }\forall a \in \mathcal{T}
\end{equation}

\paragraph{Softmax Policy} We also use a softmax policy which is an extension of the probabilistic policy. It also converts the $Q_t(a)$ values to probability distribution, however it uses a softmax function for that, 
The following equation defines the softmax policy.

\begin{equation}
\pi(a)=\frac{e^{Q_t(a)}}{\sum_{a\in\mathcal{T}}e^{Q_t(a)}} \text{,        }\forall a \in \mathcal{T}
\end{equation}

\subsubsection{Reward Modeling} In our experiments, we have used two approaches to model rewards in each epoch when a particular $TT$ value is selected for an epoch. Since the reward is based on the loss function, the first model proposes to use the inverse of the loss function as reward. Formally
\begin{equation}
    R_t(a) = \frac{1}{\mathcal{L}_t}
\label{formula:loss1}
\end{equation}

Here, $\mathcal{L}_t$ is the loss function when arm or tabu tenure $a$ was selected at epoch $t$. This reward is then fed to Equation~\ref{formula:Q} for value update. 
However, there is a problem with this approach, when the loss value is very low, then the reward will be a large value. In that case, the probability of choosing that selected $TT$ value after the  \textit{adaption period} will dominate other $TT$ values. We propose another reward model to improve over this issue. In the second model, we define the reward as the negative exponential of the loss function. Formally, it is defined as in the following:
\begin{equation}
R_t(a) = e^{-\mathcal{L}_t}
\label{formula:loss2}
\end{equation}

This function does not result into very large rewards for small losses and thus exploration is retained and allows to avoid the pitfall of the previous reward function. We have also experimented with differential loss functions, however the initial experiments did not show promising results and we did not report that in this paper. 

\section{Experimental Analysis}

We used PyTorch, a deep learning framework for quick and flexible experimentation, to implement Tabu Droput and Tabu Tenure Dropout. All the experiments were conducted on 3.60GHz Intel(R) Core(TM) i7-7700 CPU, 32 GB RAM and a NVIDIA TITAN XP with 12GB physical memory under Ubuntu 16.04.7 LIS. We run all of the algorithms five times each and on the average of the results for all of the experiments are reported in this paper.

\subsection{Implementation Details}
In this section, we provide the details of the neural network architecture and datasets used in this paper. 
\subsubsection{Neural network architectures}
For the convenience of comparison we have used the same neural network architectures as Ma et al. \cite{ma2020dropout} used in their work. We have used a multilayer perceptron (MLP) and two CNN architecture. LeNet-5, a newer version of LeNet, is the first CNN architecture \cite{lecun1989backpropagation} and the second one is named CNN1\footnote[1]{\href{https://github.com/pytorch/examples/blob/master/mnist/main.py}{https://github.com/pytorch/examples/blob/master/mnist/main.py}} as their work\cite{ma2020dropout}. LeNet5 extracts features by intelligent design using convolution, parameter sharing, pooling, and other processes, eliminating a huge amount of processing expense, and then utilizes a fully connected neural network for classification and recognition. This network has lately used as the foundation for a significant variety of neural network topologies. The architecture of LeNet5, Conv2d (Convolutional layer 1, 3 channel input, 6 convolution kernels, kernel size 5 x 5) - ReLU - MaxPool2d (2 x 2 max pooling) - Conv2d (Convolutional layer 2, 6 input channels, 16 convolution kernels, kernel size 5 x 5) - ReLU - MaxPool2d (2 x 2 max pooling) - FC - ReLU - FC - ReLU - FC - Logsoftmax.  The MLP model is a full-connected (FC) neural network with 1024 units in each hidden layer (FC-ReLU-FC-ReLU-FC-Logsoftmax). Between the two hidden layers, we introduce dropout variations.

\subsubsection{Parameter Setting}
For all of the trials, we utilized a dropout rate of 0.5. For each model, we trained 300 epochs using a learning rate of 0.01 and a batch size of 512. For all neural networks, Adam was used as the optimizer. The $TT_{max}$ value in our experiment was 6. We have used 4 \textit{adaption period} values, these are 10, 15, 20, 25.

\subsection{Datasets}
The experiments were carried out on four datasets, the specifics of which are listed below.
\begin{itemize}
    \item MNIST\footnote{\href{http://yann.lecun.com/exdb/mnist/}{http://yann.lecun.com/exdb/mnist/}}: It is a traditional handwritten digits dataset with 28 x 28 grayscale pictures of 10 digits (ranging from 0 to 9) which is frequently used in computer vision and machine learning. There are 60000 training images and 10000 images for testing. For this dataset, we have used MLP and \href{https://github.com/pytorch/examples/blob/master/mnist/main.py}{CNN-1}. Here the Figure \ref{fig:datasets} (a) shows some samples of MNIST dataset.

    \item Fashion-MNIST\footnote[3]{\href{https://github.com/zalandoresearch/fashion-mnist}{https://github.com/zalandoresearch/fashion-mnist}}: It is identical to the MNIST dataset in terms of training, test, number of class labels and image dimensions. Here is also 60000 training set, 10000 test set, 10 class labels (T-shirt/top, Trouser/pants, Pullover shirt, Dress, Coat, Sandal, Shirt, Sneaker, Bag and Ankle boot) with 28 x 28 grayscale images. Here the Figure \ref{fig:datasets} (b) shows some samples of Fashion-MNIST dataset.

    \item CIFAR-10\footnote[4]{\href{https://www.cs.toronto.edu/~kriz/cifar.html}{https://www.cs.toronto.edu/~kriz/cifar.html}}: It contains 60000 32 x 32 colored images where 50000 images for training set and 10000 for testing set. This dataset is classified into 10 different classes (airplanes, cars, birds, cats, deer, dogs, frogs, horses, ships, and trucks) and each class has 6000 images. Exactly 1000 randomly-selected images are in test set for every class but in training set there may contains more images from one class to another. Here the Figure \ref{fig:datasets} (c) shows some samples of CIFAR-10 dataset.

    \item CIFAR-100: It’s almost identical to CIFAR-10 dataset, except it has 100 classes containing 600 images each. There are also 50000 images for training set and 10000 images for test set that means 500 for training and 100 for testing per class. This 100 classes of CIFAR-100 are grouped into 20 superclasses. Every images comes with two label “fine” for class and “coarse” for superclass. Here the Figure \ref{fig:datasets} (d) shows some samples of CIFAR-100 dataset.
\end{itemize}

\subsection{Effectiveness of Tabu Tenure dropout}
The Standard Tabu dropout diversifies the neural network by dropping off the marked neurons in the successive epoch. The tabu tenure dropout further extends the tabu to a number of epochs controlled by the parameter $TT$ value. 
Table \ref{Table: Tabu Tenure} illustrates the comparison between the error rate or loss function of Standard Tabu dropout and the Tabu Tenure dropout on the testing datasets after 300 epochs. We have calculated the arithmetic mean of 5 runs to generate each value. Note that our experiments are limited to $TT=6$. The bold values in each column of the table shows the best error rate achieved using different tabu tenures. The results are narrated dataset wise in the following.  
\subsubsection{Result on different datasets} Table \ref{Table: Tabu Tenure} illustrates the result of mean error rate on different datasets. MNIST - CNN-1 obtains better performance until Tabu Tenure dropout 5. Though the Tabu Tenure dropout 6 shows a slightly higher error, the value is pretty close to the previous one. While using MLP and CNN-1, the increased value of Tabu Tenure outperforms the others. As per the fact that the dropping delay of neurons is expanding the variety of both MLP and CNN-1 gradually. Finally, the Tabu Tenure 6 dropout achieves the most favorable result among the others. We observe on the LeNet-5 model that in primary stages, the error was decreasing slowly. In the end, Tabu Tenure dropout 6 shows a significantly lower error than the  Standard Tabu dropout. The LeNet-5 model with Tabu Tenure dropout 6 has an outstanding result in all the training epochs than the Standard Tabu dropout. For all the datasets, the Tabu Tenure Dropout outperforms the state-of-the-art Standard Tabu Dropout.

\setlength\intextsep{\glueexpr\intextsep/2\relax}

\vspace*{-6.5pt}
\begin{table*}[!htb]
\setlength{\tabcolsep}{2pt}
\caption{The average error rate (after training 300 epochs) on the test datasets using Standard Tabu and Tabu Tenure dropout. The best results are marked in bold. (FMNIST is the short form of Fashion-MNIST, C10 is for CIFAR-10 and, C100 is for CIFAR-100).}
\begin{scriptsize}
\begin{tabular}{l c c c c c c} 
        \toprule
        & \multicolumn{1}{c}{MNIST - MLP} 
        & \multicolumn{1}{c}{MNIST - CNN1}
        & \multicolumn{1}{c}{FMNIST - MLP}
        & \multicolumn{1}{c}{FMNIST - CNN1} 
        & \multicolumn{1}{c}{C10 - LeNet5} 
        & \multicolumn{1}{c}{C100 - LeNet5}\\
        \midrule
        AlphaDropout          & 0.01162 & 0.00602 & 0.11989 & 0.03800 & 0.14851 & 0.74702 \\
        Non-dropout           & 0.01302 & 0.00729 & 0.12471 & 0.03856 & 0.15707 & 0.74258 \\
        Standard dropout      & 0.01076 & 0.00582 & 0.11242 & 0.03754 & 0.13904 & 0.73983 \\
        Standard Tabu Dropout & 0.01083 & 0.00585 & 0.11886 & 0.03731 & 0.13744 & 0.73408 \\
        Tabu Tenure Dropout 1 & 0.00989 & 0.00625 & 0.11639 & 0.03765 & 0.14418 & 0.71176 \\
        Tabu Tenure Dropout 2 & 0.00811 & 0.00459 & 0.09762 & 0.03073 & 0.11643 & 0.48707 \\
        Tabu Tenure Dropout 3 & 0.00689 & 0.00316 & 0.08587 & 0.02312 & 0.10501 & 0.38576 \\
        Tabu Tenure Dropout 4 & 0.00768 & 0.00351 & 0.07747 & 0.02208 & 0.09340 & 0.34184 \\
        Tabu Tenure Dropout 5 & 0.00749 & \textbf{0.00306} & 0.07269 & 0.01936 & 0.08589 & 0.29113\\
        Tabu Tenure Dropout 6 & \textbf{0.00675} & 0.00317 & \textbf{0.06733} & \textbf{0.01885} & \textbf{0.08169} & \textbf{0.27844} \\

        \bottomrule
        \label{Table: Tabu Tenure}
\end{tabular}%
\end{scriptsize}
\end{table*}

\begin{table*}[t!]
  \setlength{\tabcolsep}{2pt}
  \caption{The mean error rate (after training 300 epochs) on the test datasets using different arm choosing methods for Adaptive Tabu Dropout. (TD is the short form of Tabu Dropout, AP is for \textit{Adaption Period}, SP is for Softmax Probabilistic, FMNIST is the short form of Fashion-MNIST, C10 is for CIFAR-10 and, C100 is for CIFAR-100. V1 refers to Equation \ref{formula:loss1} and V2 refers to Equation \ref{formula:loss2}).}
  \begin{scriptsize}
    \begin{tabular}{lrrrrrr}
    \toprule
          & \multicolumn{1}{c}{MNIST - MLP} & \multicolumn{1}{c}{MNIST - CNN1} & \multicolumn{1}{c}{FMNIST - MLP} & \multicolumn{1}{c}{FMNIST - CNN1} & \multicolumn{1}{c}{C10 - LeNet5} & \multicolumn{1}{c}{C100 - LeNet5} \\
    \midrule
        TD Random (AP: 10) & 0.00886 & 0.00372 & 0.08983 & 0.02759 & 0.11402 & 0.43774 \\
    TD Random (AP: 15) & 0.00849 & 0.00385 & 0.08689 & 0.02489 & 0.10563 & 0.43635 \\
    TD Random (AP: 20) & 0.00782 & 0.00332 & 0.08902 & 0.02533 & 0.11064 & 0.44873 \\
    TD Random (AP: 25) & 0.00769 & 0.00394 & 0.08936 & 0.02597 & 0.10686 & 0.47377 \\
    \midrule
    TD Greedy (AP: 10) & 0.00924 & 0.00369 & 0.08399 & 0.02271 & 0.09782 & 0.38999 \\
    TD Greedy (AP: 15) & 0.00701 & 0.00409 & 0.07995 & 0.02257 & 0.10278 & 0.39555 \\
    TD Greedy (AP: 20) & 0.00749 & 0.00374 & 0.08365 & 0.02509 & 0.10275 & 0.38272 \\
    TD Greedy (AP: 25) & 0.00851 & 0.00439 & 0.09014 & 0.02318 & 0.10397 & 0.42020 \\
    \midrule
    TD Epsilon Greedy (AP: 10) & 0.00824 & 0.00402 & 0.08504 & 0.02415 & 0.10667 & 0.41873 \\
    TD Epsilon Greedy (AP: 15) & 0.00777 & 0.00414 & 0.08455 & 0.02163 & 0.10751 & 0.37835 \\
    TD Epsilon Greedy (AP: 20) & 0.00759 & 0.00358 & 0.07988 & 0.02269 & 0.10184 & 0.45175 \\
    TD Epsilon Greedy (AP: 25) & 0.00787 & 0.00392 & 0.08630 & 0.02232 & 0.09759 & 0.39445 \\
    \midrule
    TD Probabilistic (AP: 10) & 0.00766 & 0.00336 & 0.08013 & \textbf{0.01958} & 0.09592 & 0.36828 \\
    TD Probabilistic (AP: 15) & 0.00747 & \textbf{0.00307} & 0.09351 & 0.02248 & 0.10738 & 0.38269 \\
    TD Probabilistic (AP: 20) & 0.00756 & 0.00339 & 0.08649 & 0.02370 & 0.09672 & 0.39072 \\
    TD Probabilistic (AP: 25) & 0.00754 & 0.00360 & 0.08706 & 0.02005 & 0.09730 & 0.38556 \\
    \midrule
    TD SP (AP: 10) V1 & 0.00763 & 0.00398 & 0.08199 & 0.02630 & 0.10471 & 0.33584 \\
    TD SP (AP: 15) V1 & 0.00661 & 0.00378 & 0.07691 & 0.02558 & 0.09352 & 0.36606 \\
    TD SP (AP: 20) V1 & 0.00791 & 0.00361 & 0.10153 & 0.02356 & 0.10184 & 0.38102 \\
    TD SP (AP: 25) V1 & 0.00775 & 0.00349 & 0.07419 & 0.02211 & 0.10017 & 0.35419 \\
    \midrule
    TD SP (AP: 10) V2 & 0.00766 & 0.00332 & \textbf{0.06729} & 0.02493 & \textbf{0.08722} & 0.30933 \\
    TD SP (AP: 15) V2 & 0.00753 & 0.00321 & 0.06913 & 0.02559 & 0.09969 & 0.29918 \\
    TD SP (AP: 20) V2 & 0.00707 & 0.00392 & 0.06827 & 0.02497 & 0.09236 & \textbf{0.29515} \\
    TD SP (AP: 25) V2 & \textbf{0.00651} & 0.00386 & 0.06969 & 0.02144 & 0.09419 & 0.30444 \\
    \bottomrule
    \end{tabular}%
    \end{scriptsize}
  \label{Table: Adaptive Tabu}%
\end{table*}%



\subsection{Effectiveness of Adaptive Tabu dropout} 
We have formulated the Adaptive Tabu dropout as a multi-armed bandit problem where the arms of the bandit are the $TT$ values selected from a set $\mathcal{T}$. We have used a set of policies: random, greedy, epsilon greedy, probabilistic, and softmax probabilistic techniques for analysis. Two different reward models have been used. The Tabu Tenure values are changed after  \textit{adaption periods}. In the experiments we have used four different values of adaption periods (AP) of 10, 15, 20, and 25 epochs. Table \ref{Table: Adaptive Tabu} shows the error rate of various tabu tenure selection policies with different adaption periods. Each value reported in Table \ref{Table: Adaptive Tabu} is average of 5 runs. In case of the softmax policy, we have used two types of reward models defined by Equation~\ref{formula:loss1} and Equation~\ref{formula:loss2}. Other policies are experimented with only with the reward model defined in Equation~\ref{formula:loss1}. The bold values in each column represents the best values achieved by any policy. 

From the values reported in the table, we note that softmax policy based tabu tenure selection achieves best values among 5 out of 6 combinations of dataset-architectures. In the case of MNIST dataset with CNN1 architecture, probabilistic policy performes better than softmax policy. Note that the greedy policy does not work well. In case of greedy we allowed an initial phase for exploration (150 epochs) and then followed the greedy policy. Results indicate that explorations should be encouraged in all stages of training. 


If we compare the performances of adaptive selection or policy based tabu tenure with that of the fixed tabu tenure methods, we note that the results are very much similar. However, to test them further we have used AlexNet\footnote{\href{https://github.com/Lornatang/pytorch-alexnet-cifar100/blob/master/model.py}{https://github.com/Lornatang/pytorch-alexnet-cifar100/blob/master/model.py}} architecture \cite{krizhevsky2012imagenet} which is one of the earliest successful deep neural network on the relatively larger datasets CIFAR-10 and CIFAR-100. 

We have considered the average accuracy and error of the best-fixed tabu tenure version ($TT=6$) and the best policy-based algorithm softmax policy with different adaptation periods. We have examined the average accuracy and error of the best-fixed tabu tenure version ($TT=6$) and the best policy-based algorithm softmax policy with different adaptation periods. Softmax policy with adaption period 10 shows the minimum error on both CIFAR-10 and CIFAR-100 datasets. The highest accuracy is observed on Softmax policy with an adaption period of 25 on the CIFAR-10 dataset.  Note that all four combinations of dataset-architecture policy-based adaptive tabu tenure with softmax policy are achieving significantly superior values compared to the fixed tabu tenure version. This encourages us to conclude on the overall effectiveness of the adaptive tabu selection method.

\subsection{Comparison with other methods}
Note the We have reported the experimental results in Table \ref{Table: Tabu Tenure} on the identical four datasets after training with 300 epochs using same neural network models as done in \cite{ma2020dropout}. In their experiments, Ma et al. \cite{ma2020dropout} showed the superiority of their tabu dropout over AlphaDropout \cite{klambauer2017self}, Curriculum Dropout \cite{morerio2017curriculum}, Standard Dropout\cite{hinton2012improving,krizhevsky2012imagenet}, and non-dropout strategies. Table \ref{Table: Tabu Tenure} implies a clear contrast with the other dropout approaches and shows that the Tabu Tenure Dropout outperforms all the dropout strategies mentioned in \cite{ma2020dropout}.  
The Adaptive Tabu Tenure is very much similar in performance compared to the fixed tabu tenure method and outperforms fixed tabu tenure in terms of error and accuracy using AlexNet architecture on larger datasets. Therefore, we conclude that the Adaptive Tabu Tenure Dropout outperforms the rest of the dropout techniques with whom \cite{ma2020dropout} has shown comparison in their paper. 

\section{Conclusion}
In this paper, we have introduced adaptive tabu dropout based on the multi-arm bandit problem which is a simple mechanism that can diversify the neural network more and perform better compared to the standard Tabu dropout and Tabu Tenure dropout. While using Tabu Tenure dropout we need to select ta $TT$ value wisely otherwise there may happen an overfitting problem. In our experiments, we have noticed that $TT$ values beyond 6 overfits and also larger $TT$ values run for longer epochs results in overfitting. This encourages to use multiple tabu tenures and adaptive selection policy. Also note that we have not explored much of the differential reward models and non-stationary modeling of the rewards for a better estimation of the $Q_t(a)$ values. The effectiveness of the methods proposed in this paper encourages us to assess Adaptive Tabu dropout in larger and differnt types of deep neural network models with longer training periods, such as RNN, Res-Net, etc and bigger datasets, such as ImageNet \cite{deng2009imagenet}.

\bibliographystyle{splncs04}
\bibliography{references.bib}
\end{document}